\documentclass{article}
\PassOptionsToPackage{numbers, compress}{natbib}
\usepackage[preprint]{neurips_2025}
\usepackage[utf8]{inputenc} % allow utf-8 input
\usepackage[T1]{fontenc}    % use 8-bit T1 fonts
\usepackage[colorlinks=true, allcolors=Blue]{hyperref}       % hyperlinks
\usepackage{url}            % simple URL typesetting
\usepackage{booktabs}       % professional-quality tables
\usepackage{amsfonts}       % blackboard math symbols
\usepackage{nicefrac}       % compact symbols for 1/2, etc.
\usepackage{microtype}      % microtypography
\usepackage[table,dvipsnames]{xcolor}         % colors
\usepackage{wrapfig}
\usepackage[size=small]{caption}
\usepackage{mathtools}
\usepackage{amssymb}
\usepackage{amsthm}
\usepackage[algoruled,boxed,lined,noend]{algorithm2e}
\usepackage{adjustbox} 
\usepackage{float}
\usepackage{amssymb}% http://ctan.org/pkg/amssymb
\usepackage{pifont}% http://ctan.org/pkg/pifont
\usepackage{subcaption}
\usepackage{todonotes}
\usepackage{lineno}
\usepackage{enumitem}
\usepackage{bm}
\usepackage{chngcntr}
\usepackage{svg}
\usepackage{todonotes}
\renewcommand*{\thefootnote}{\fnsymbol{footnote}}
\usepackage{authblk}
\usepackage{placeins}

\definecolor{amplifai}{HTML}{230f87}
\newcommand{\prompt}[1]{{\color{amplifai}\noindent #1}}

\title{AMPLIFAI: A Multiphase CT Dataset for Benchmarking Clinical Reasoning in LI-RADS Assessment of Liver Lesions}

\begin{document}
\author[$*$1,2]{\textbf{Pranav Kulkarni}}
\author[$*$1,4]{\textbf{Nikhil Shah}}
\author[1,2]{\textbf{Amritansh Suryavanshi}}
\author[1]{\textbf{Jana G. Delfino}}
\author[3]{\textbf{James Tonascia}}
\author[3]{\textbf{Jade Wong-You-Cheong}}
\author[3]{\textbf{Barton Lane}}
\author[5]{\textbf{Joseph Chirico}}
\author[3]{\textbf{Jeffrey D. Hirsch}}
\author[1,3]{\textbf{Ang Li}}
\author[1,2]{\textbf{Heng Huang}}
\author[$\dagger$1,3]{\textbf{Florence X. Doo}}

\affil[1]{University of Maryland Institute for Health Computing, North Bethesda, MD 20852}
\affil[2]{Department of Computer Science, University of Maryland, College Park, MD 20742}
\affil[3]{Department of Electrical and Computer Engineering, University of Maryland, College Park, MD 20742}
\affil[4]{Department of Diagnostic Radiology and Nuclear Medicine, University of Maryland School of Medicine, Baltimore, MD 21201}
\affil[5]{University of Maryland Medical System, Baltimore, MD 21201}

\renewcommand{\thefootnote}{\fnsymbol{footnote}}
\footnotetext[1]{These authors contributed equally to this work.}
\footnotetext[2]{Corresponding author(s): \texttt{fdoo@som.umaryland.edu}}
\maketitle

\begin{abstract}
Hepatocellular carcinoma (HCC) is the third leading cause of cancer-related mortality worldwide, with early detection improving survival from <20\% to >70\%. The standardized Liver Imaging Reporting and Data System (LI-RADS) criteria provide an imaging-based diagnostic framework to evaluate liver lesions for HCC, serving as a foundation for automating HCC detection with artificial intelligence (AI). However, the lack of large, publicly available datasets with high-quality annotations has limited the development and evaluation of AI models for automated LI-RADS assessment. We introduce \textbf{AMPLIFAI} dataset, the first public dataset of 590 multiphase abdominal CT studies annotated with LI-RADS categories, lesion size, and voxel-level segmentations for three major LI-RADS features: arterial phase hyperenhancement, washout, and enhancing capsule. The dataset was curated and harmonized from four public datasets and augmented with expert annotations from five board-certified radiologists and one resident. Following the \emph{Datasheets for Datasets} format, this paper details the dataset's composition, curation and harmonization process, and annotation workflow to support transparent, reproducible research in medical imaging AI.
\end{abstract}
\section{Introduction} \label{sec:introduction}

Hepatocellular carcinoma (HCC) is the most common primary liver cancer and the third leading cause of cancer-related mortality worldwide, where early detection can improve patient survival from <20\% up to >70\% \cite{singal2014early,llovet2024author}. The standardized Liver Imaging Reporting and Data System (LI-RADS) \cite{chernyak2018liver,lee2023percentages} defines the diagnostic criteria to evaluate liver lesions for HCC, relying on multiphase abdominal CT as one of the primary imaging modalities. Specifically, LI-RADS category 5 (LR-5) is considered "definitely HCC," \cite{goins2023conversion,goins2023individual,adamo2025diagnostic} establishing a biopsy-free, fully imaging-based diagnostic pathway where artificial intelligence (AI) could enable early detection and improve patient outcomes \cite{laino2022added,ying2024multicenter,mule2023automated}.

\begin{figure}[!t]
    \centering
    \includegraphics[width=\linewidth]{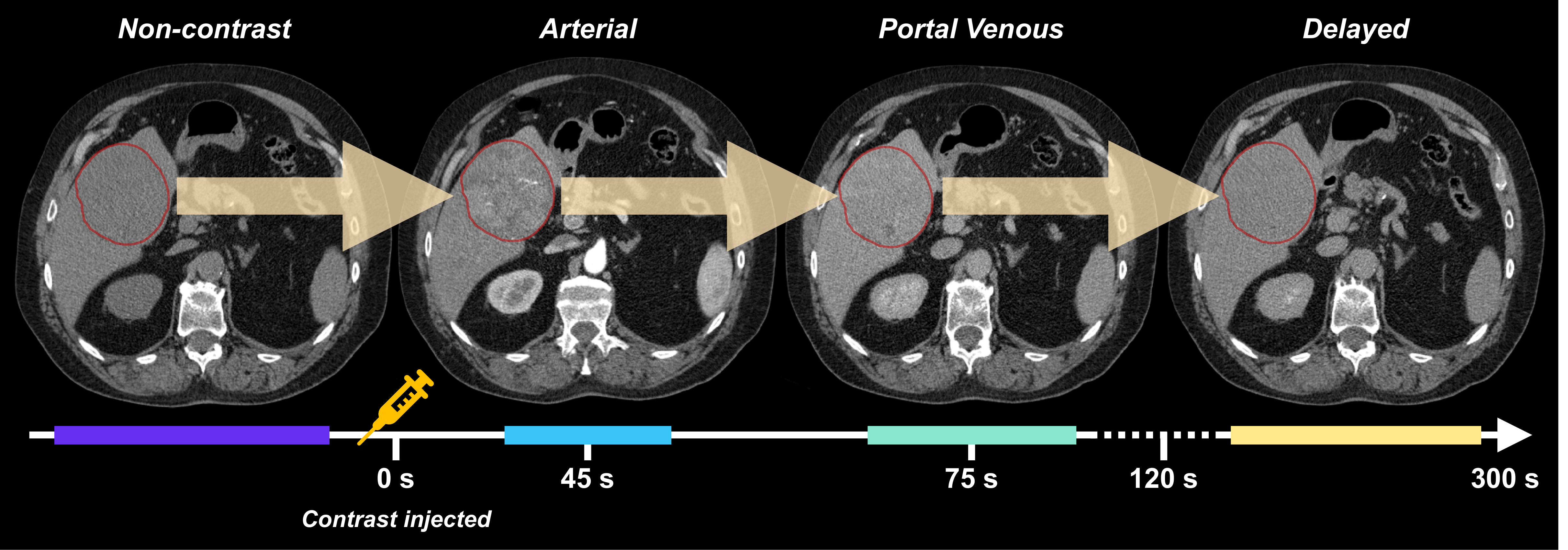}
    \caption{Dynamic contrast enhancement patterns of liver lesions in multiphase abdominal CT before contrast injection (non-contrast) and at successive time points after injection: arterial, portal venous, and delayed phases.}
    \label{fig:contrast_phases}
\end{figure}

However, automated LI-RADS assessment presents a unique challenge that distinguishes it from other medical imaging classification tasks \cite{laino2022added}. LI-RADS assessment relies on multiphase abdominal CT acquisitions to capture the dynamic enhancement patterns of liver lesions (Figure \ref{fig:contrast_phases}). Consequently, AI models must account for both phase-specific lesion characteristics and cross-phase feature interactions. This complexity makes data curation highly time-consuming, as scans need to be spatially registered and annotated for features across multiple phases. As a result, there is a distinct lack of large, publicly available multiphase CT datasets with granular, feature-level annotations of the major LI-RADS features required to develop and evaluate AI models for automated LI-RADS assessment of liver lesions.

Here, we introduce \textbf{AMPLIFAI} (\textbf{A}nnotated \textbf{M}ulti\textbf{P}hase \textbf{L}iver \textbf{I}maging \textbf{F}or \textbf{AI}), a benchmark dataset for LI-RADS assessment of liver lesions in multiphase abdominal CT. AMPLIFAI consists of 590 cases curated and harmonized from four public datasets \cite{erickson2016the,bartnik2024waw,moawad2023multimodality,luo2025comprehensive}. We augmented these cases with expert annotations from five board-certified radiologists and one resident, including LI-RADS categories (LR-1 through LR-5, LR-TIV, and LR-M), lesion size (in mm), and voxel-level segmentation masks for three major imaging features: non-rim arterial phase hyperenhancement (APHE), non-peripheral washout, and enhancing capsule. Our dataset establishes the first multiphase CT dataset with high-quality, reliable ground-truth for benchmarking LI-RADS assessment of liver lesions.

In this paper, we provide structured documentation of AMPLIFAI following the \emph{Datasheets for Datasets} format \cite{gebru2021datasheets}. We describe the dataset composition, curation and harmonization pipeline, and annotation workflow to support transparent and reproducible research in medical imaging AI as per the FAIR Data Principles \cite{wilkinson2016fair}. This datasheet only covers the publicly released training and validation data as part of the \href{https://um2ii.github.io/amplifai-challenge/}{AMPLIFAI Challenge Proposal} at MICCAI 2026 Satellite Events. While the held-out private institutional test set will not be publicly released to maintain the integrity of the challenge, it was preprocessed and annotated using the same procedures as the publicly released data.

\section{AMPLIFAI Datasheet}
\subsection{Motivation} \label{sec:datasheet_motivation}

\prompt{\textbf{For what purpose was the dataset created?} Was there a specific task in mind? Was there a specific gap that needed to be filled? Please provide a description.}

\begin{figure}[!t]
    \centering
    \includegraphics[width=0.8\linewidth]{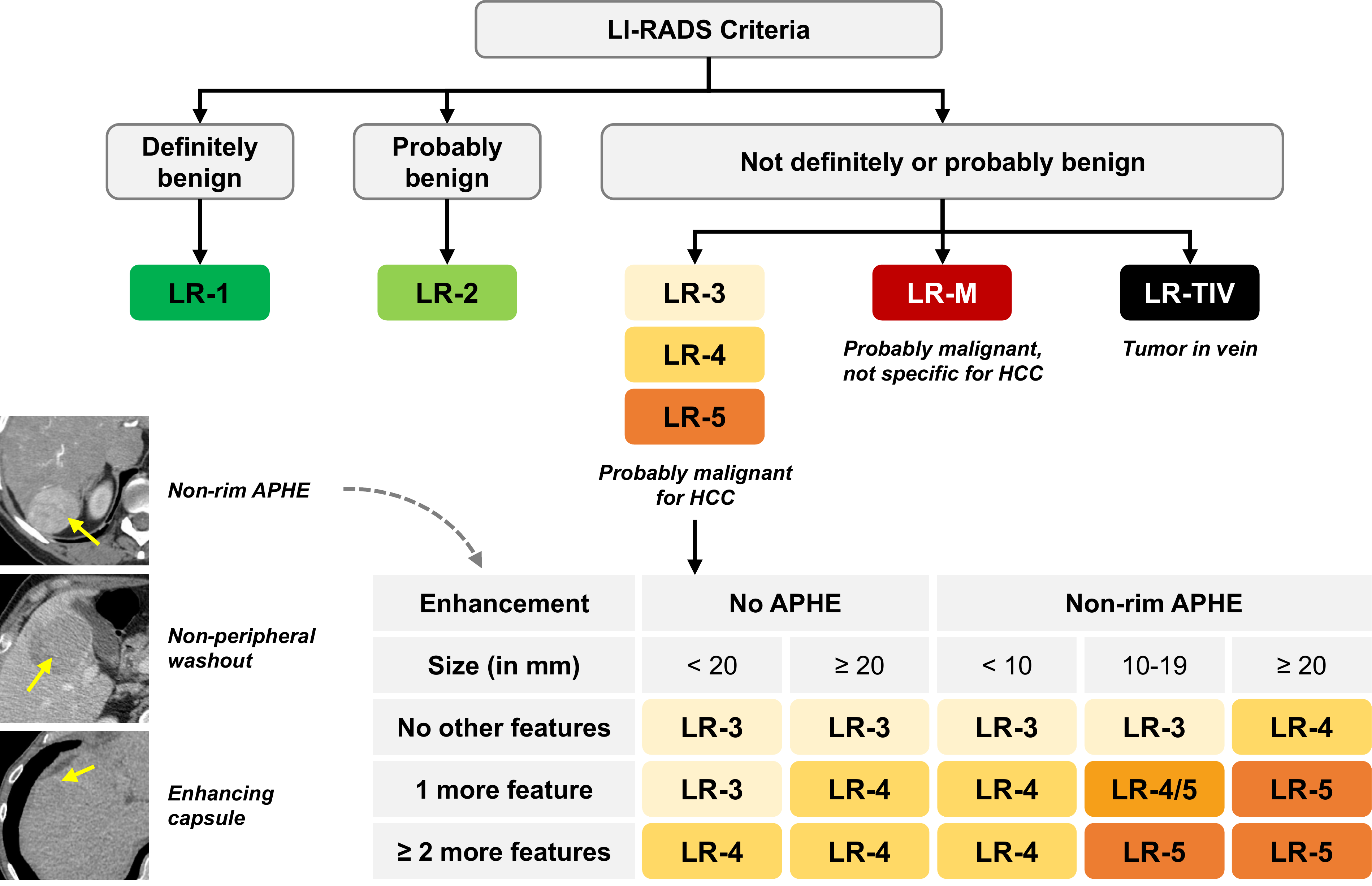}
    \caption{An overview of the standardized LI-RADS criteria\protect\footnotemark[1]. Categories LR-1 to LR-5 describe the increasing probability of HCC. LR-M indicates a lesion that is malignant but not specific for HCC, while LR-TIV denotes definite tumor in vein. Major LI-RADS features include lesion size (in mm), non-rim arterial phase hyperenhancement (APHE), non-peripheral washout, and enhancing capsule.}
    \label{fig:criteria}
\end{figure}

AMPLIFAI was created to support the development and evaluation of artificial intelligence (AI) models for automated assessment of liver lesions in multiphase abdominal CT using the standardized LI-RADS criteria. Specifically, the dataset addresses the lack of large, publicly available CT datasets with expert annotations for both LI-RADS categories and major imaging features that the criteria relies upon.

The LI-RADS criteria provide a standardized diagnostic framework for evaluating liver lesions based on their characteristics and dynamic contrast enhancement patterns \cite{chernyak2018liver}. For patients at risk for HCC, categories LR-1 through LR-5 describe the increasing probability of HCC: LR-1 indicates a lesion is "definitely benign", while LR-5 is considered "definitely HCC" and does not require further tissue biopsy for diagnosis. In addition, LI-RADS includes LR-M for lesions that are probably or definitely malignant but have features that are not specific to HCC, and LR-TIV for the presence of unequivocal tumor in vein. These categories are clinically important because they distinguish HCC from malignant features that may or may be associated with HCC.

The LI-RADS criteria relies on the temporal analysis of liver lesions in multiphase CT, where multiple contrast phases are acquired: before contrast injection (non-contrast) and at successive time points after injection (arterial, portal venous, and delayed) to capture dynamic enhancement patterns in the lesion (Figure \ref{fig:criteria}). The primary gating feature is arterial phase hyperenhancement (APHE), where the lesion appears brighter than the surrounding healthy tissue in the early arterial phase. It is considered "non-rim" if the enhancement spreads internally and is not limited to the outer edge of the lesion. Only lesions exhibiting unequivocal non-rim APHE are considered for the highest probability categories (LR-3 to LR-5). However, if the enhancement is limited to the outer edge, the lesion indicates rim APHE and cannot be considered for LR-5.
\footnotetext[1]{Figure inspired by Radiology Assistant: https://radiologyassistant.nl/abdomen/liver/li-rads}

If APHE is present, the LI-RADS category is determined by a combination of the lesion's size and how APHE interacts with two major features: First, non-peripheral washout, where the lesion becomes darker than the liver in the portal venous or delayed phases. Second, enhancing capsule, where a thin, peripheral rim remains enhanced in the portal venous or delayed phases due to contrast retention. The lesion size and presence of one or both features increases the probability for HCC. For example, a lesion with non-RIM APHE that is 2.3cm and has one additional feature (washout or capsule) is categories as LR-5 ("definitely HCC") while a lesion without non-rim APHE can only reach a maximum of LR-4 ("probably HCC"), even if it is large and shows washout.

\begin{figure}[!t]
    \centering
    \includegraphics[width=\linewidth]{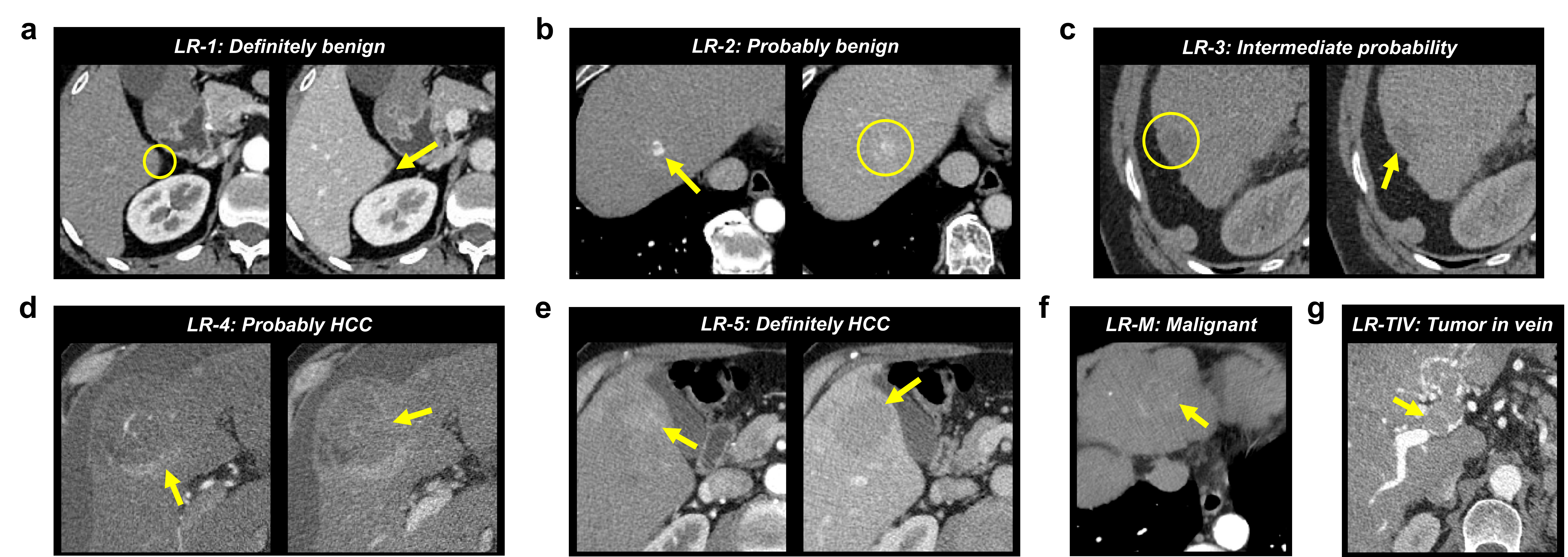}
    \caption{Representative examples from each LI-RADS category. \textbf{(a)} The LR-1 image displays a lesion with no enhancement. \textbf{(b)} The LR-2 image shows a small (10 mm) lesion with non-rim APHE but no malignant features. \textbf{(c)} The LR-3 image shows an intermediate (30 mm) lesion with unequivocal malignant features. \textbf{(d)} The LR-4 image shows a large (93 mm) lesion with non-rim APHE. \textbf{(e)} The LR-5 image shows a large (63 mm) lesion with both non-rim APHE and non-peripheral washout. \textbf{(f)} The LR-M image shows a large (30 mm) lesion with rim APHE but lacks malignant features specific for HCC. \textbf{(g)} the LR-TIV image shows a large (51 mm) lesion with soft tissue present in the vein.}
    \label{fig:examples}
\end{figure}

This presents a unique challenge for automating LI-RADS assessment of liver lesions. AI models must capture the spatio-temporal relationships between lesion morphology and dynamic enhancement patterns across multiple contrast phases (Figure \ref{fig:examples}). Despite the strong potential to enable early diagnosis and improve patient outcomes, there are no large, publicly available datasets with granular, radiologist-annotated labels for major LI-RADS features to support the development and evaluation of AI-based LI-RADS assessment.

AMPLIFAI addresses this gap as the first public dataset to provide LI-RADS categories (LR-1 through LR-5, LR-M, and LR-TIV), lesion size (in mm), and voxel-level segmentation masks for the three major imaging features: non-rim APHE, non-peripheral washout, and enhancing capsule. The dataset was curated and harmonized from four public datasets and augmented with annotations from five board-certified radiologists and one resident. By providing both LI-RADS categories and feature-level segmentation masks, AMPLIFAI establishes a benchmark to support the development and evaluation of AI models for automated LI-RADS assessment in multiphase CT.

\prompt{\textbf{Who created the dataset (e.g., which team, research group) and on behalf of which entity (e.g., company, institution, organization)?}}

AMPLIFAI was created by the \href{https://ihc.umd.edu/research-centers/applied-ai/}{Center for Applied AI} team at the \href{https://ihc.umd.edu/}{University of Maryland Institute for Health Computing (UM-IHC)}. Our interdisciplinary team includes researchers, faculty, and clinicians from the University of Maryland, College Park, the University of Maryland School of Medicine, and the University of Maryland Medical System. The complete list of authors is provided on the title page.

\prompt{\textbf{Who funded the creation of the dataset?} If there is an associated grant, please provide the name of the grantor and the grant name and number.}

The UM-IHC Center for Applied AI team is supported in part by the Montgomery County, Maryland and the University of Maryland Strategic Partnership: MPowering the State, a formal collaboration between the University of Maryland, College Park and the University of Maryland, Baltimore. We also acknowledge the support of the 2026 Accelerated Translational Incubator Pilot (ATIP) Grant Program from the University of Maryland, Baltimore, Institute for Clinical and Translational Research (ICTR), which is funded in part by the National Center for Advancing Translational Sciences (NCATS) Clinical Translational Science Award (CTSA), UM1TR004926. F.D. is supported in part by a grant NIH CTSA 1K12TR004925-01A1. The contents of this manuscript are solely the responsibility of the authors and do not necessarily represent the official view of the National Institutes of Health.

\subsection{Composition} \label{sec:datasheet_composition}

\prompt{\textbf{What do the instances that comprise the dataset represent (e.g., documents, photos, people, countries)?} Are there multiple types of instances (e.g., movies, users, and ratings; people and interactions between them; nodes and edges)? Please provide a description.}

Each instance in the dataset represents a single medical imaging examination of a patient evaluated for HCC using the standardized LI-RADS criteria.

\prompt{\textbf{How many instances are there in total (of each type, if appropriate)?}}

The dataset contains 590 cases from 584 unique patients, derived from four public datasets: TCGA-LIHC \cite{erickson2016the}, WAW-TACE \cite{bartnik2024waw}, HCC-TACE-SEG \cite{moawad2023multimodality}, and PLC-CECT \cite{luo2025comprehensive}. Most instances correspond to a single patient, with the exception of TCGA-LIHC \cite{erickson2016the}, which includes multiple visits (up to 3) for 5 patients. To prevent data leakage, we split the dataset into a training set of 531 cases (90\%) and validation set of 59 cases (10\%) without any patient overlap. Uniform sampling was used across the source datasets to ensure a balanced data distribution in both sets. Table \ref{tab:splits} shows the total number of cases in the training and validation sets, stratified by their source datasets. 

The majority of the dataset (about $85\%$) is derived from PLC-CECT and WAW-TACE \cite{luo2025comprehensive,bartnik2024waw}. These well-documented and standardized datasets have large sample sizes and require minimal preprocessing and harmonization. In contrast, the HCC-TACE-SEG and TCGA-LIHC \cite{moawad2023multimodality,erickson2016the} datasets have greater heterogeneity, requiring extensive preprocessing prior to inclusion. Section \ref{sec:datasheet_preprocessing} provides a detailed description of our data curation pipeline.

\begin{table}[!t]
    \small
    \caption{Distribution of cases across source datasets for training and validation splits.}
    \label{tab:splits}
    \setlength{\tabcolsep}{12pt}
    \centering
    \begin{tabular}{lccc} \toprule
        \textbf{Source} & \textbf{Training} & \textbf{Validation} & \textbf{Total} \\ \midrule
        PLC-CECT \cite{luo2025comprehensive} & $277$ ($52.17\%$) & $29$ ($49.15\%$) & $306$ ($51.86\%$) \\
        WAW-TACE \cite{bartnik2024waw} & $176$ ($33.15\%$) & $23$ ($38.98\%$) & $199$ ($33.73\%$) \\ 
        HCC-TACE-SEG \cite{moawad2023multimodality} & $41$ ($7.72\%$) & $4$ ($6.78\%$) & $45$ ($7.63\%$) \\ 
        TCGA-LIHC \cite{erickson2016the} & $37$ ($6.97\%$) & $3$ ($5.08\%$) & $40$ $(6.78\%$) \\ \midrule
        \textbf{Total} & $531$ ($100\%$) & $59$ ($100\%$) & $590$ ($100\%$) \\ \bottomrule
    \end{tabular}
\end{table}

\begin{table}[!t]
    \small
    \caption{Distribution of contrast phases across training and validation splits.}
    \label{tab:phase}
    \setlength{\tabcolsep}{12pt}
    \centering
    \begin{tabular}{lcccc} \toprule
        \textbf{Phase} & \textbf{Non-contrast} & \textbf{Arterial} & \textbf{Portal Venous} & \textbf{Delayed} \\ \midrule
        Training & $381$ ($71.75\%$) & $531$ ($100\%$) & $527$ ($99.25\%$) & $406$ ($76.46\%$) \\
        Validation & $43$ ($72.88\%$) & $59$ ($100\%$) & $59$ ($100\%$) & $45$ ($76.27\%$) \\ \midrule
        \textbf{Total} & $424$ ($71.86\%$) & $590$ ($100\%$) & $586$ ($99.32\%$) & $451$ ($76.44\%$) \\ \bottomrule
    \end{tabular}
\end{table}
    
\prompt{\textbf{Does the dataset contain all possible instances or is it a sample (not necessarily random) of instances from a larger set?} If the dataset is a sample, then what is the larger set? Is the sample representative of the larger set (e.g., geographic coverage)? If so, please describe how this representativeness was validated/verified. If it is not representative of the larger set, please describe why not (e.g., to cover a more diverse range of instances, because instances were withheld or unavailable).}

The dataset represents a non-random sample of studies from four public datasets: TCGA-LIHC was collected at the Mayo Clinic, the University of North Carolina, Alberta Health Services, and Lahey Hospital and Medical Center \cite{erickson2016the}; WAW-TACE was collected at the Medical University of Warsaw between May 2016 and April 2021 \cite{bartnik2024waw}; HCC-TACE-SEG was collected at the University of Texas MD Anderson Cancer Center between November 2002 and June 2012 \cite{moawad2023multimodality}; and PLC-CECT was collected at the Chongqing Yubei District People's Hospital between January 2015 and December 2022 \cite{luo2025comprehensive}. We only included multiphase abdominal CT studies of patients prior to HCC treatment in this analysis. Please refer to each source dataset's publication for further details.

\prompt{\textbf{What data does each instance consist of?} “Raw” data (e.g., unprocessed text or images) or features? In either case, please provide a description.}

Each instance consists of a multiphase contrast-enhanced abdominal CT study, containing one arterial and at least one late contrast phase (portal venous and/or delayed). Non-contrast scans are included when available. Table \ref{tab:phase} shows the distribution of contrast phases in the dataset. We provide all CT volumes and segmentation masks in the NIfTI format for accessibility.

\prompt{\textbf{Is there a label or target associated with each instance?} If so, please provide a description.}

Each case contains annotations for the LI-RADS category, lesion size (in mm), and the presence of three major LI-RADS features: non-rim APHE, non-peripheral washout, and enhancing capsule. Since patients may present with multiple liver lesions, we define a \emph{target lesion} for each case as the lesion considered most suspicious for HCC, and therefore, most relevant for LI-RADS assessment. For cases with multiple lesions, the target lesion is selected based on the annotator's judgment of the most probable lesion for HCC. Cases without a liver lesion were also included in the dataset as normal cases. 

\begin{figure}[!t]
    \centering
    \includegraphics[width=0.9\linewidth]{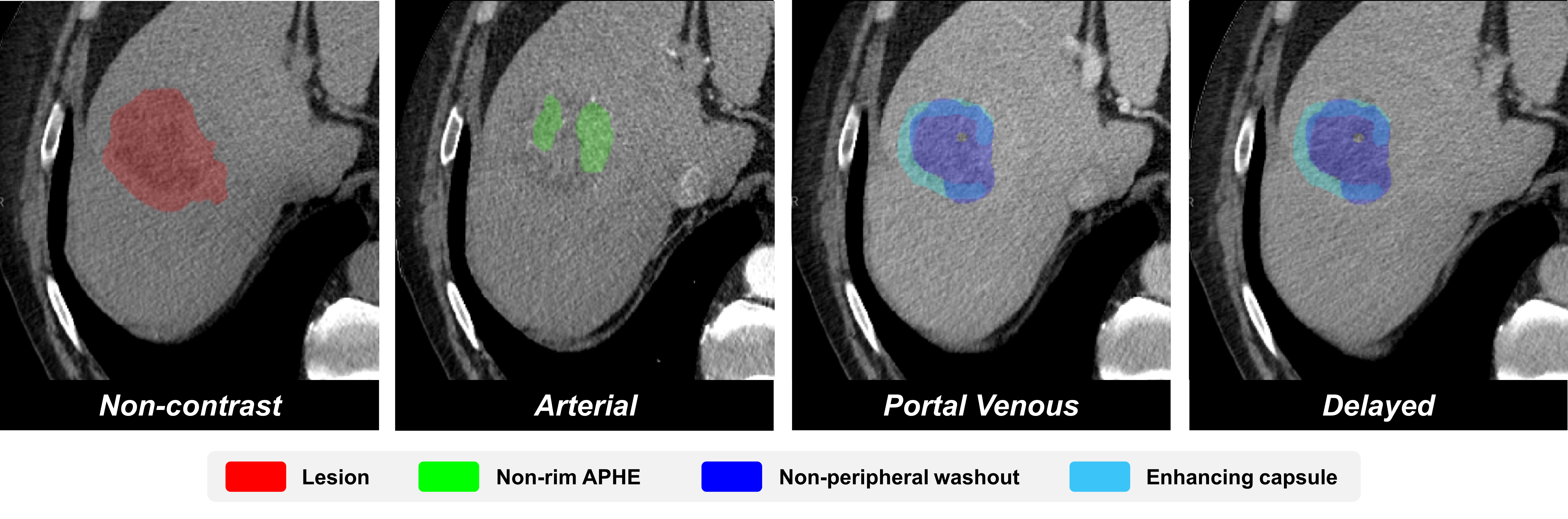}
    \caption{A multiphase abdominal CT with segmentation masks for the target lesion and three major LI-RADS features: arterial phase hyperenhancement (APHE), non-peripheral washout, and enhancing capsule.}
    \label{fig:segmentation}
\end{figure}

For each target lesion, we provide a voxel-level segmentation mask of the lesion, and, when applicable, segmentation masks for the three major LI-RADS imaging features (Figure \ref{fig:segmentation}). Importantly, since non-peripheral washout and enhancing capsule may be present in either or both late-contrast phases, we provide phase-specific segmentation masks for these features in every phase in which they are present. These target lesions are labeled with one of seven standardized LI-RADS categories \cite{chernyak2018liver}:
\begin{itemize}[leftmargin=2em]
    \item LR-1 ("Definitely benign"): Lesions show no enhancement in any contrast phase.
    \item LR-2 ("Probably benign"): Small, distinct nodules without the presence of major features.
    \item LR-3 ("Intermediate probability"): Lesions with equivocal presence of major features.
    \item LR-4 ("Probably HCC"): Highly suspicious lesions depending on their size and the presence of some major features.
    \item LR-5 ("Definitely HCC"): Large lesions with unequivocal presence of major features.
    \item LR-TIV ("Tumor in vein"): Unequivocal presence of soft tissue within a vein, regardless of whether an associated lesion is observed.
    \item LR-M ("Malignant"): Lesions that appear malignant but lack the typical characteristics of HCC.
\end{itemize}

Note that for the purposes of this challenge, LI-RADS categorization was determined strictly using major LI-RADS features. In standard clinical practice, "ancillary features" are a natural part of the LI-RADS algorithm and routinely used by radiologists to upgrade or downgrade a lesion's category. However, to focus the scope of this challenge, final LI-RADS categorization was determined strictly using major features; ancillary features were not evaluated or utilized. For cases with a target lesion, lesion size is measured as the longest dimension (in mm) of the target lesion in the axial plane. The presence of major LI-RADS features is provided as a binary label (i.e., 0/1), with the exception of APHE, which distinguishes between non-rim APHE, rim APHE, and the absence of APHE. All labels are included in the metadata CSV file of the training and validation splits.

\prompt{\textbf{Is any information missing from individual instances?} If so, please provide a description, explaining why this information is missing (e.g., because it was unavailable). This does not include intentionally removed information, but might include, e.g., redacted text.}

All instances are complete.

\prompt{\textbf{Are relationships between individual instances made explicit (e.g., users' movie ratings, social network links)?} If so, please describe how these relationships are made explicit.}

There are no known relationships between the patients.

\prompt{\textbf{Are there recommended data splits (e.g., training, development/validation, testing)?} If so, please provide a description of these splits, explaining the rationale behind them.}

We provide training and validation splits, containing $531$ ($90\%$) and $59$ ($10\%$) of cases. These are derived through uniform sampling of the resulting dataset to ensure a balanced distribution of labels in both sets. In addition, we prevented any patient overlap between training and validation sets from multi-visit examinations in TCGA-LIHC \cite{erickson2016the}. Table \ref{tab:label_splits} shows the distribution of annotations across training and validation splits.

\begin{table}[!t]
    \small
    \caption{Distribution of annotations across training and validation splits.}
    \label{tab:label_splits}
    \setlength{\tabcolsep}{12pt}
    \centering
    \begin{tabular}{lccc} \toprule
        \textbf{Variables} & \textbf{Training} & \textbf{Validation} & \textbf{Total} \\ \midrule
        Lesion size (in mm) & $57.83 \pm 43.31$ & $56.81 \pm 36.98$ & $57.72 \pm 42.79$ \\
        \textbf{APHE} \\
            \quad Non-rim APHE & $258$ ($48.59\%$) & $33$ ($55.93\%$) & $291$ ($49.32\%$) \\
            \quad Rim APHE & $62$ ($11.68\%$) & $7$ ($11.86\%$) & $69$ ($11.69\%$) \\
            \quad Absent & $211$ ($39.73\%$) & $19$ ($32.20\%$) & $230$ ($38.98\%$) \\
        \textbf{Washout} \\ 
            \quad Venous & $185$ ($34.84\%$) & $24$ ($40.68\%$) & $209$ ($35.42\%$) \\
            \quad Delayed & $178$ ($33.52\%$) & $19$ ($32.20\%$) & $197$ ($33.39\%$) \\
        \textbf{Capsule} \\ 
            \quad Venous & $84$ ($15.82\%$) & $14$ ($23.73\%$) & $98$ ($16.61\%$) \\
            \quad Delayed & $59$ ($11.11\%$) & $4$ ($6.78\%$) & $63$ ($10.68\%$) \\
        \textbf{LI-RADS category} \\ 
            \quad LR-1 & $2$ ($0.38\%$) & $1$ ($1.69\%$) & $3$ ($0.51\%$) \\
            \quad LR-2 & $2$ ($0.38\%$) & $1$ ($1.69\%$) & $3$ ($0.51\%$) \\
            \quad LR-3 & $13$ ($2.45\%$) & $3$ ($5.08\%$) & $16$ ($2.71\%$) \\
            \quad LR-4 & $41$ ($7.72\%$) & $6$ ($10.17\%$) & $47$ ($7.97\%$) \\
            \quad LR-5 & $231$ ($43.50\%$) & $27$ ($45.76\%$) & $258$ ($43.73\%$) \\
            \quad LR-TIV & $57$ ($10.73\%$) & $7$ ($11.86\%$) & $64$ ($10.85\%$) \\
            \quad LR-M & $115$ ($21.66\%$) & $14$ ($23.73\%$) & $129$ ($21.86\%$) \\ \bottomrule
    \end{tabular}
\end{table}

\prompt{\textbf{Are there any errors, sources of noise, or redundancies in the dataset?} If so, please provide a description.}

Potential sources of noise, variability, and error in the dataset stem from both clinical heterogeneity and technical data-curation challenges. These include:
\begin{itemize}[leftmargin=2em]
    \item \textbf{Inter-annotator variability:} Expert interpretation and boundary delineation naturally vary; prior literature indicates radiologist segmentation performance can fluctuate by 10-30\% \cite{van2025measurement}

    \item \textbf{Image quality and artifacts:} Standard CT artifacts (e.g., metallic streak artifacts, patient motion, or low signal-to-noise ratio) are present in a subset of cases, which may result in suboptimal diagnostic quality. 
    
    \item \textbf{Spatial registration and preprocessing}: Due to respiratory motion between contrast phase acquisitions, some multiphase series exhibit residual spatial registration errors despite computational alignment efforts. Additionally, a small number of cases suffered from the loss of DICOM affine transformation matrix metadata during anonymization or format conversion; where recoverable, orientation was aligned manually, but severe cases were excluded during quality control.

    \item \textbf{Protocol incompleteness:} While complete multiphase acquisitions (non-contrast, arterial, portal venous, and delayed phases) were targeted, some historical cases had incomplete phase protocols (e.g., missing a delayed or non-contrast phase, or limited to single/dual-phase studies). These were retained where diagnostic evaluation of the target lesion remained clinically feasible, reflecting real-world clinical workflow heterogeneity.

    \item \textbf{Contrast phase timing heterogeneity:} Variation in contrast bolus timing across scan acquisitions (e.g., early vs. late arterial phase) introduces variability in the visual manifestation of arterial phase hyperenhancement (APHE).
    
    \item \textbf{Reconstruction parameters and voxel anisotropy:} Heterogeneity in slice thickness and reconstruction kernels across different scanner vendors introduces partial volume effects and voxel spacing anisotropy. This can make thin features, such as an enhancing capsule, more challenging to segment.
    
    \item \textbf{Co-occurring pathologies:} Background parenchymal noise resulting from underlying severe cirrhosis, diffuse steatosis (fatty liver disease), or previous locoregional therapies can complicate feature isolation.
    
    \item \textbf{Class imbalance:} Natural class imbalance exists within the dataset. Because the data originates from a tertiary care clinical cohort, there is a higher prevalence of high-risk lesions (e.g., LR-4 and LR-5) compared to strictly benign (LR-1/LR-2) or rare categories (LR-TIV).
    
\end{itemize}

\prompt{\textbf{Is the dataset self-contained, or does it link to or otherwise rely on external resources (e.g., websites, tweets, other datasets)?} If it links to or relies on external resources, a) are there guarantees that they will exist, and remain constant, over time; b) are there official archival versions of the complete dataset (i.e., including the external resources as they existed at the time the dataset was created); c) are there any restrictions (e.g., licenses, fees) associated with any of the external resources that might apply to a dataset consumer? Please provide descriptions of all external resources and any restrictions associated with them, as well as links or other access points, as appropriate.}

The dataset is self-contained.

\prompt{\textbf{Does the dataset contain data that might be considered confidential (e.g., data that is protected by legal privilege or by doctor-patient confidentiality, data that includes the content of individuals' non-public communications)?} If so, please provide a description.}

All data is anonymized and personally identifiable information has been removed. Please refer to each source dataset's publication for further details.

\prompt{\textbf{Does the dataset contain data that, if viewed directly, might be offensive, insulting, threatening, or might otherwise cause anxiety?} If so, please describe why.}

No.

\prompt{\textbf{Does the dataset identify any subpopulations (e.g., by age, gender)?} If so, please describe how these subpopulations are identified and provide a description of their respective distributions within the dataset.}

No.

\prompt{\textbf{Is it possible to identify individuals (i.e., one or more natural persons), either directly or indirectly (i.e., in combination with other data) from the dataset?} If so, please describe how.}

No.

\prompt{\textbf{Does the dataset contain data that might be considered sensitive in any way (e.g., data that reveals race or ethnic origins, sexual orientations, religious beliefs, political opinions or union memberships, or locations; financial or health data; biometric or genetic data; forms of government identification, such as social security numbers; criminal history)?} If so, please provide a description.}

No.
\subsection{Collection Process} \label{sec:datasheet_collection}

\begin{figure}[!t]
    \centering
    \includegraphics[width=0.9\linewidth]{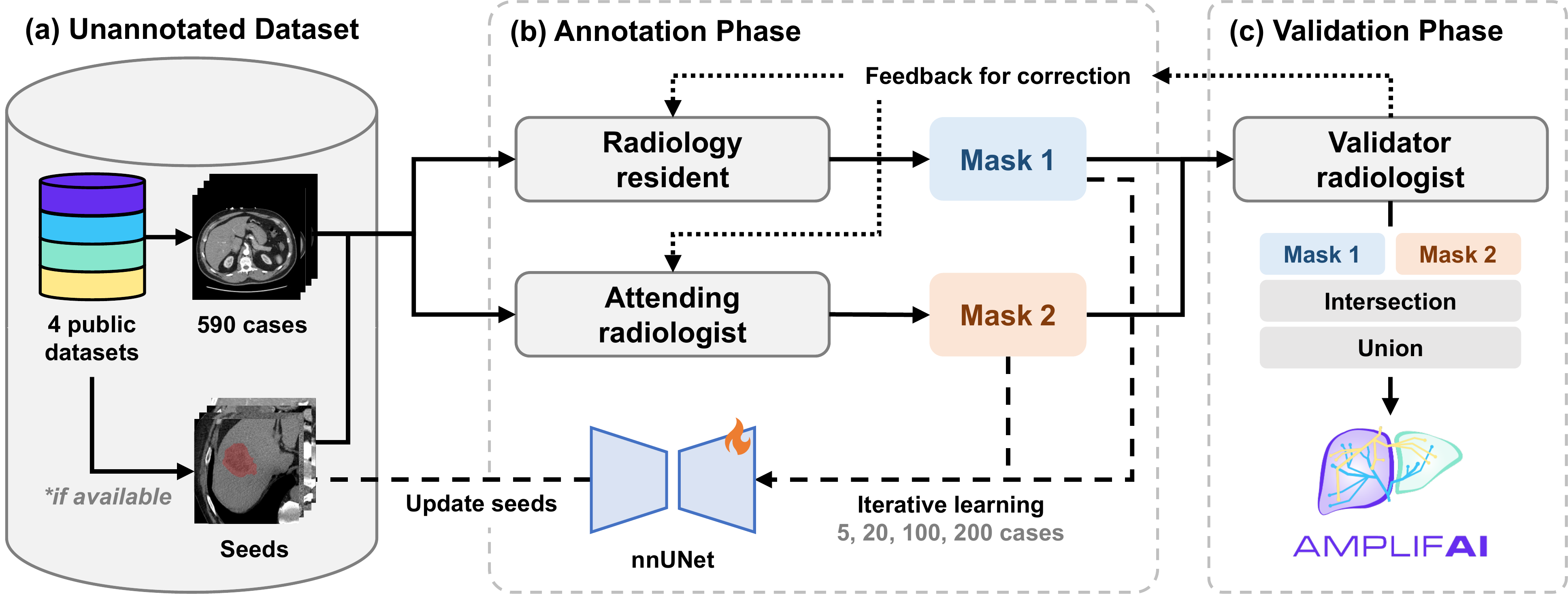}
    \caption{Overview of the annotation workflow. \textbf{(a)} We curated and harmonized 590 cases from four public datasets, using available liver lesion segmentations as initial seeds. \textbf{(b)} Each case was independently annotated by a radiology resident and one of three attending radiologists, including LI-RADS category, lesion size, and segmentation of three major LI-RADS features. An iteratively trained nnU-Net was used to update segmentation seeds throughout annotation process. \textbf{(c)} Cases with two independent annotations were reviewed by one of two independent validators, who selected or combined the annotations. Cases with disagreements were returned to the annotators for correction before final validation.}
    \label{fig:annotation_workflow}
\end{figure}

\prompt{\textbf{How was the data associated with each instance acquired?} Was the data directly observable (e.g., raw text, movie ratings), reported by subjects (e.g., survey responses), or indirectly inferred/derived from other data (e.g., part-of-speech tags, model-based guesses for age or language)? If the data was reported by subjects or indirectly inferred/derived from other data, was the data validated/verified? If so, please describe how.}

The dataset was completed by a team of five board-certified radiologists and one radiology resident with substantial clinical experience in abdominal imaging. Four attendings (J.T., J.W., B.L., F.D.) and the resident (J.C.) served as primary annotators, with validation conducted by two attendings (J.H., F.D.). The attending who served as both an annotator and a validator (F.D.) was restricted from and did not validate her own annotated cases. The annotation team was onboarded to the workflow and attended regular check-in meetings to maintain progress. They were blinded to the dataset splits, and the annotation protocol remained consistent for every case.

Each case was first independently annotated by the resident and one attending annotator. They were provided with annotations for liver lesions (if the source dataset provided them) as well as AI-generated segmentation masks for each of the major LI-RADS features. We followed an iterative annotation workflow \cite{wasserthal2023totalsegmentator}: after manual segmentation of the first five cases was completed, we trained a preliminary nnU-Net \cite{isensee2021nnu}, and its predictions acted as seeds for manual refinement in order to reduce annotation time. The nnU-Net was retrained once new annotations were available at 5, 20, 100, and 200 cases, with a completely new model trained at each step to generate new predictions. Thus, cases assigned for annotation after each retraining step received seeds generated by the newly trained model, while cases that had already been annotated were not re-annotated with the updated model. The annotators were blinded to this iterative learning process and were not informed of which model version generated the provided seeds.

Once a case was annotated by the resident and an attending annotator, it proceeded further for validation. Importantly, the validators were not involved in the initial annotations and did not directly adjudicate disagreements. They were provided with the option to pick between either one of the two annotations, or their union or intersection. In the case of disagreement (e.g., equivocal features or annotators selecting different target lesions), the validators could sent back the initial annotations to the respective annotators with feedback to verify and address the disagreement. The annotators then reviewed the case and made any necessary corrections before the case proceeded through validation again. Regarding the imaging data acquisition, please refer to each source dataset's publication for further details.

\prompt{\textbf{What mechanisms or procedures were used to collect the data (e.g., hardware apparatuses or sensors, manual human curation, software programs, software APIs)?} How were these mechanisms or procedures validated?}

\begin{figure}[!t]
    \centering
    \includegraphics[width=\linewidth]{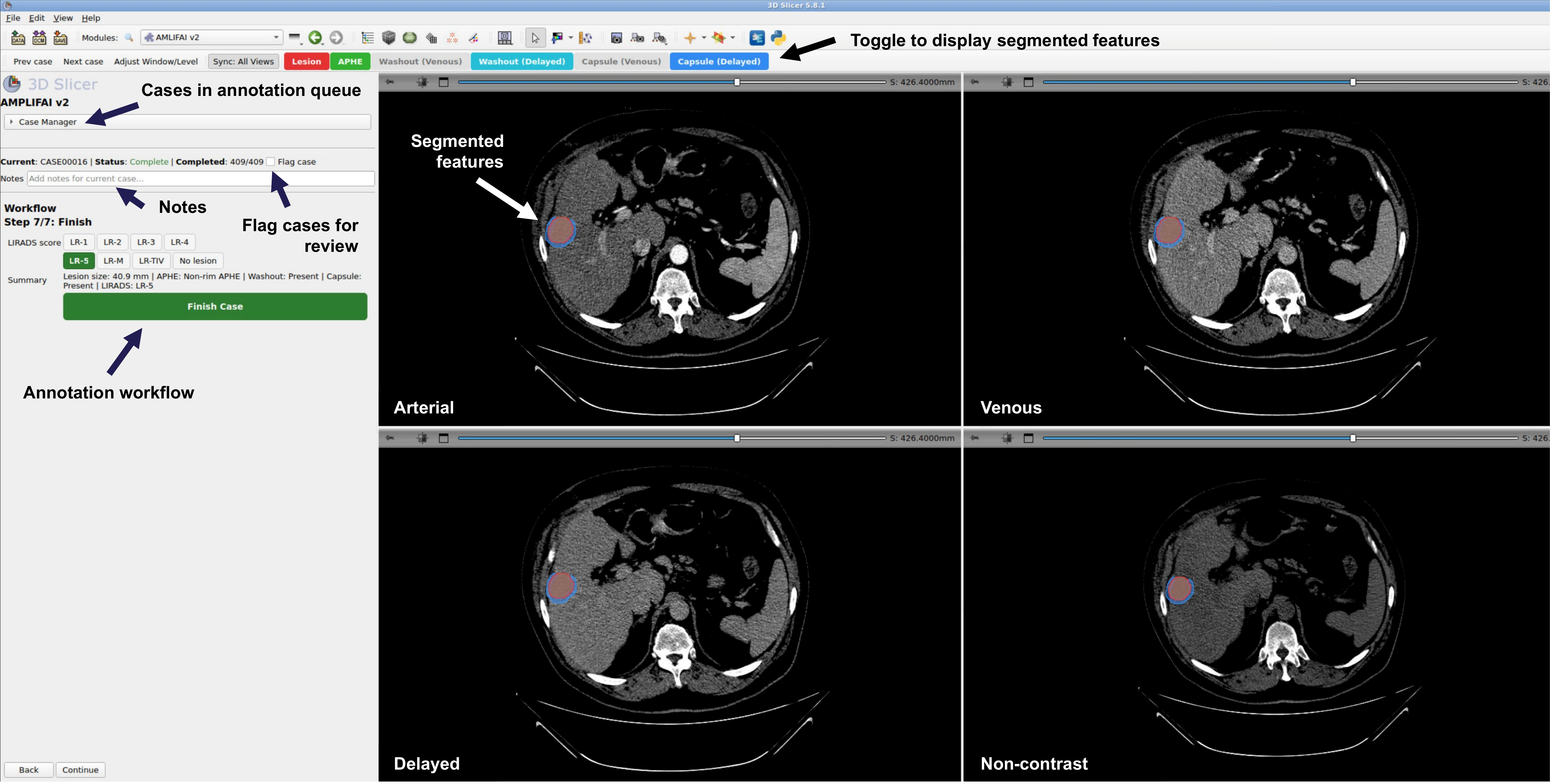}
    \caption{Overview of the custom 3D Slicer module for the radiologist annotation workflow.}
    \label{fig:slicer_ui}
\end{figure}

We developed a custom 3D Slicer module to support both the annotation and validation workflows (Figure \ref{fig:slicer_ui}). It was hosted in a secure research environment located at our hospital. All primary annotations and validations were performed within this environment, with the radiologists remotely accessing the module using authentication that followed institutional protocols.

\prompt{\textbf{If the dataset is a sample from a larger set, what was the sampling strategy (e.g., deterministic, probabilistic with specific sampling probabilities)?}}

No.

\prompt{\textbf{Who was involved in the data collection process (e.g., students, crowdworkers, contractors) and how were they compensated (e.g., how much were crowdworkers paid)?}}

An interdisciplinary team from the \href{https://ihc.umd.edu/research-centers/applied-ai/}{Center for Applied AI (CA2i)} at the \href{https://ihc.umd.edu/}{University of Maryland Institute for Health Computing (UM-IHC)} curated the dataset. Our team includes researchers, faculty, and clinicians from the University of Maryland, College Park, the University of Maryland School of Medicine, and the University of Maryland Medical System.
 
The annotation team, affiliated with the Department of Diagnostic Radiology and Nuclear Medicine within the School of Medicine, University of Maryland Baltimore, consisted of five board-certified radiologists and a radiology resident with substantial clinical experience in abdominal imaging at the time of this project (2026): Florence X. Doo (8 years of experience); Jeffrey D. Hirsch (25 years of experience); Barton F. Lane (24 years of experience); James Tonascia (6 years of experience); Jade Wong-You-Cheong (39 years of experience); and Joseph Chirico (4th year diagnostic radiology resident).
 
Compensation for the clinical annotation team was provided through Dr. Ang Li's and Dr. Florence X. Doo's research funds, supported in part by the ICTR ATIP grant. Please refer to the Section \ref{sec:datasheet_motivation} for full grant details.

\prompt{\textbf{Over what timeframe was the data collected?} Does this timeframe match the creation timeframe of the data associated with the instances (e.g., recent crawl of old news articles)? If not, please describe the timeframe in which the data associated with the instances was created.}

The dataset consists of multiphase CT studies from four public datasets collected between November 2002 and December 2022. Please refer to each source dataset's publication for further details. The annotations were performed between May 2026 and June 2026.

\prompt{\textbf{Were any ethical review processes conducted (e.g., by an institutional review board)?} If so, please provide a description of these review processes, including the outcomes, as well as a link or other access point to any supporting documentation.}

An IRB exemption was granted to conduct this retrospective study by the University of Maryland Institutional Review Board (HP-00117149).

\prompt{\textbf{Does the dataset relate to people?} If not, you may skip the remaining questions in this section.}

Yes, the dataset contains abdominal CT scans from human patients.

\prompt{\textbf{Did you collect the data from the individuals in question directly, or obtain it via third parties or other sources (e.g., websites)?}}

The dataset is derived from four public datasets: TCGA-LIHC \cite{erickson2016the}, WAW-TACE \cite{bartnik2024waw}, HCC-TACE-SEG \cite{moawad2023multimodality}, and PLC-CECT \cite{luo2025comprehensive}. Please refer to each source dataset's publication for further details.

\prompt{\textbf{Were the individuals in question notified about the data collection?} If so, please describe (or show with screenshots or other information) how notice was provided, and provide a link or other access point to, or otherwise reproduce, the exact language of the notification itself.}

Patients were not directly notified since the data is sourced from publicly available repositories. Please refer to each source dataset's publication for further details.

\prompt{\textbf{Did the individuals in question consent to the collection and use of their data?} If so, please describe (or show with screenshots or other information) how consent was requested and provided, and provide a link or other access point to, or otherwise reproduce, the exact language to which the individuals consented.}

Please refer to each source dataset's publication for further details.

\prompt{\textbf{If consent was obtained, were the consenting individuals provided with a mechanism to revoke their consent in the future or for certain uses?} If so, please provide a description, as well as a link or other access point to the mechanism (if appropriate)}

There is no mechanism to remove a patient from the dataset since the data is sourced from publicly available repositories. Please refer to each source dataset's publication for further details.

\prompt{\textbf{Has an analysis of the potential impact of the dataset and its use on data subjects (e.g., a data protection impact analysis) been conducted?} If so, please provide a description of this analysis, including the outcomes, as well as a link or other access point to any supporting documentation.}

All data is anonymized and contains no personally identifiable information. Please refer to each source dataset's publication for further details.
\subsection{Preprocessing/Cleaning/Labeling} \label{sec:datasheet_preprocessing}

\prompt{\textbf{Was any preprocessing/cleaning/labeling of the data done (e.g., discretization or bucketing, tokenization, part-of-speech tagging, SIFT feature extraction, removal of instances, processing of missing values)?} If so, please provide a description. If not, you may skip the remaining questions in this section.}

The following preprocessing and harmonization steps were applied to standardize the imaging data across all four source datasets. The inclusion and exclusion criteria were applied consistently across the datasets, resulting in the final dataset of 590 cases.

\begin{enumerate}
    \item \textbf{Source dataset:} We began with 796 cases across the four source datasets: WAW-TACE \cite{bartnik2024waw} ($n=233$), PLC-CECT \cite{luo2025comprehensive} ($n=361$), HCC-TACE-SEG \cite{moawad2023multimodality} ($n=105$), and TCGA-LIHC \cite{erickson2016the} ($n=97$, multi-visit cases). We removed irrelevant DICOM and DICOM-SEG series that were not suitable for our annotation workflow. For example, TCGA-LIHC \cite{erickson2016the} contained irrelevant series such as Scout and Dose Report series, while HCC-TACE-SEG \cite{moawad2023multimodality} contained combined lesion segmentations that were not compatible with our target lesion-based annotation workflow. After these initial exclusions and dataset-specific filtering, 230 cases proceeded further from HCC-TACE-SEG and TCGA-LIHC \cite{moawad2023multimodality,erickson2016the}.

    \item \textbf{Conversion to NIfTI:} TCGA-LIHC \cite{erickson2016the} and HCC-TACE-SEG \cite{moawad2023multimodality} were provided in DICOM format and were converted to NIfTI using dcm2niix\footnote{https://github.com/rordenlab/dcm2niix} for standardization across the dataset. This conversion also improves accessibility of the imaging data for further research use. WAW-TACE \cite{bartnik2024waw} provided lesion segmentations in NRRD format, which were converted to NIfTI to maintain format consistency across all segmentation masks.

    \item \textbf{Contrast phase detection:} We classified the contrast phase of each volume as non-contrast, arterial, portal venous, or delayed, first using available imaging metadata. If the study lacked sufficient metadata for phase detection, we used Comp2Comp \cite{wasserthal2023totalsegmentator,blankemeier2023comp2comp} to predict the contrast phase. Phase detection was used to determine whether a case met our inclusion criteria: eligible cases were required an arterial phase and at least one late-contrast phase (portal venous or delayed). Cases that did not meet these requirements were excluded. After phase detection, only 140 cases from HCC-TACE-SEG and TCGA-LIHC \cite{moawad2023multimodality,erickson2016the} were eligible for further processing

    \item \textbf{Image registration:} All included multiphase CT studies were registered with ANTsPy\footnote{https://github.com/antsx/antspy} to a fixed reference phase. The portal venous phase was selected as the primary reference. If the portal venous phase was not present, the delayed phase was used instead. Registration was performed to spatially align the available contrast phases and enable consistent voxel-level segmentation across phases.

    \item \textbf{Registration quality:} To identify poorly registered studies, we computed the Dice score of rigid anatomical structures (i.e., bones) segmented with TotalSegmentator \cite{wasserthal2023totalsegmentator} between the registered and reference phases. A registration was considered acceptable when the Dice score was above an empirically determined threshold of $0.64$. Cases that did not meet this threshold were excluded. This resulted in $n=591$ cases passing the strict registration-quality criterion across all four source datasets.

    \item \textbf{Harmonization:} Following registration, the remaining cases from all four source datasets were combined and harmonized. We then excluded one case because it contained duplicate venous phase, resulting in a final dataset of $n=590$ cases: 199 from WAW-TACE \cite{bartnik2024waw}, 306 from PLC-CECT \cite{luo2025comprehensive}, 45 from HCC-TACE-SEG \cite{moawad2023multimodality}, and 40 from TCGA-LIHC \cite{erickson2016the}.

    \item \textbf{Post-annotation harmonization:} Once a case was independently annotated and validated, the assigned LI-RADS category was re-computed based on the established diagnostic framework for consistency.
\end{enumerate}

The complete preprocessing code will be made publicly available for reproducibility.

\prompt{\textbf{Was the "raw" data saved in addition to the preprocessed/cleaned/labeled data (e.g., to support unanticipated future uses)?} If so, please provide a link or other access point to the "raw" data.}

The raw data is publicly accessible through each source dataset's repository. \href{https://www.cancerimagingarchive.net/collection/tcga-lihc/}{TCGA-LIHC} and \href{https://www.cancerimagingarchive.net/collection/hcc-tace-seg/}{HCC-TAGE-SEG} are both hosted on The Cancer Imaging Archive (TCIA), \href{https://zenodo.org/records/12741586}{WAW-TACE} is hosted on Zenodo, and \href{https://www.scidb.cn/en/detail?dataSetId=d685a0b9f8974a2a9d7c880be1dc36e9}{PLC-CECT} is hosted on ScienceDB.

\prompt{\textbf{Is the software that was used to preprocess/clean/label the data available?} If so, please provide a link or other access point.}

The complete preprocessing code will be made publicly available to ensure transparency and reproducibility.
\subsection{Uses} \label{sec:datasheet_uses}

\prompt{\textbf{Has the dataset been used for any tasks already?} If so, please provide a description.}

The dataset has been used to train several models for characterizing HCC lesions using the standardized LI-RADS criteria. These models take multiphase abdominal CT scans and a target lesion mask as input to predict the probability of its associated LI-RADS category (LR-1 through LR-5, LR-TIV, or LR-M). We explore two possible methods: First, an interpretable model may strictly follow the clinical criteria to predict all major LI-RADS features (tumor size, non-rim APHE, non-peripheral washout, and enhancing capsule) before deriving the final LI-RADS category. Second, an end-to-end "black-box" model may directly predict the LI-RADS category, using either all available contrast phases, or, in some cases, only a single phase.

\prompt{\textbf{Is there a repository that links to any or all papers or systems that use the dataset?} If so, please provide a link or other access point.}

We will include a partial list of papers that use the dataset on the \href{https://um-ihc-ca2i.github.io/amplifai-challenge/}{AMPLIFAI Challenge Website}.

\prompt{\textbf{What (other) tasks could the dataset be used for?}}

The dataset can be used for several tasks beyond LI-RADS characterization. A potential application is improving the performance of medical foundation models for visual-question answering \cite{chen2026deeptumorvqa} and reasoning \cite{li2026radthinking}.

\prompt{\textbf{Is there anything about the composition of the dataset or the way it was collected and preprocessed/cleaned/labeled that might impact future uses?} For example, is there anything that a dataset consumer might need to know to avoid uses that could result in unfair treatment of individuals or groups (e.g., stereotyping, quality of service issues) or other risks or harms (e.g., legal risks, financial harms)? If so, please provide a description. Is there anything a dataset consumer could do to mitigate these risks or harms?}

No.

\prompt{\textbf{Are there tasks for which the dataset should not be used?} If so, please provide a description.}

The dataset should only be used for personal, non-commercial research. The dataset has neither been reviewed nor approved by the U.S. Food and Drug Administration (FDA) for clinical use. It should not be used for attempting to identify any individual nor diagnose patients with pathologies.
\subsection{Distribution} \label{sec:datasheet_distribution}

\prompt{\textbf{Will the dataset be distributed to third parties outside of the entity (e.g., company, institution, organization) on behalf of which the dataset was created?} If so, please provide a description.}

The dataset is publicly available on the \href{https://um-ihc-ca2i.github.io/amplifai-challenge/}{AMPLIFAI Challenge} website.

\prompt{\textbf{How will the dataset will be distributed (e.g., tarball on website, API, GitHub)?} Does the dataset have a digital object identifier (DOI)?}

The dataset is distributed to challenge participants via \href{https://huggingface.co/datasets/UM-IHC-CA2i/AMPLIFAI}{Hugging Face}, containing all images, labels, and metadata for the training and validation splits. The total dataset size is about 146 GB.

\prompt{\textbf{When will the dataset be distributed?}}

The dataset is currently available.

\prompt{\textbf{Will the dataset be distributed under a copyright or other intellectual property (IP) license, and/or under applicable terms of use (ToU)?} If so, please describe this license and/or ToU, and provide a link or other access point to, or otherwise reproduce, any relevant licensing terms or ToU, as well as any fees associated with these restrictions.}

The dataset is released under \href{https://creativecommons.org/licenses/by-nc-sa/4.0/deed.en}{CC BY-NC-SA} license for personal, non-commercial research use.

\prompt{\textbf{Have any third parties imposed IP-based or other restrictions on the data associated with the instances?} If so, please describe these restrictions, and provide a link or other access point to, or otherwise reproduce, any relevant licensing terms, as well as any fees associated with these restrictions.}

No.

\prompt{\textbf{Do any export controls or other regulatory restrictions apply to the dataset or to individual instances?} If so, please describe these restrictions, and provide a link or other access point to, or otherwise reproduce, any supporting documentation.}

No. All data is sourced from publicly available repositories.
\subsection{Maintenance} \label{sec:datasheet_maintenance}

\prompt{\textbf{Who will be supporting/hosting/maintaining the dataset?}}

The dataset will be maintained by University of Maryland Institute for Health Computing.

\prompt{\textbf{How can the owner/curator/manager of the dataset be contacted (e.g., email address)?}}

Please contact Dr. Florence X. Doo via email: \texttt{fdoo@som.umaryland.edu}. For any other general concerns, please contact: \texttt{amplifai@som.umaryland.edu}.

\prompt{\textbf{Is there an erratum?} If so, please provide a link or other access point.}

No.

\prompt{\textbf{Will the dataset be updated (e.g., to correct labeling errors, add new instances, delete instances)?} If so, please describe how often, by whom, and how updates will be communicated to dataset consumers (e.g., mailing list, GitHub)?}

We may expand the dataset for future iterations of the AMPLIFAI Challenge. Any data corrections will be posted on the \href{https://um-ihc-ca2i.github.io/amplifai-challenge/}{AMPLIFAI Challenge} website and \href{https://huggingface.co/datasets/UM-IHC-CA2i/AMPLIFAI}{Hugging Face}.

\prompt{\textbf{If the dataset relates to people, are there applicable limits on the retention of the data associated with the instances (e.g., were the individuals in question told that their data would be retained for a fixed period of time and then deleted)?} If so, please describe these limits and explain how they will be enforced.}

There are no limits on the retention of the information in the dataset.

\prompt{\textbf{Will older versions of the dataset continue to be supported/hosted/maintained?} If so, please describe how. If not, please describe how its obsolescence will be communicated to dataset consumers.}

No.

\prompt{\textbf{If others want to extend/augment/build on/contribute to the dataset, is there a mechanism for them to do so?} If so, please provide a description. Will these contributions be validated/verified? If so, please describe how. If not, why not? Is there a process for communicating/distributing these contributions to dataset consumers? If so, please provide a description.}

No. Since the data is released under \href{https://creativecommons.org/licenses/by-nc-sa/4.0/deed.en}{CC BY-NC-SA}, users are encouraged to extend, augment, or build on the dataset for non-commercial, research use \emph{with attribution}. We encourage users to reach out to us if they are planning to expand or augment the dataset so we can potentially collaborate or participate in the effort.

\bibliographystyle{unsrt}
\bibliography{refs}

\end{document}